# CHANGE OF WORD TYPES TO WORD TOKENS RATIO IN THE COURSE OF TRANSLATION (BASED ON RUSSIAN TRANSLATIONS OF K. VONNEGUT'S NOVELS)

Andrey Kutuzov (akutuzov72@gmail.com), Tyumen State University

*Abstract*

*The article provides lexical statistical analysis of K. Vonnegut's two novels and their Russian translations. It is found out that there happen some changes between the speed of word types and word tokens ratio change in the source and target texts. The author hypothesizes that these changes are typical for English-Russian translations, and moreover, they represent an example of Baker's translation feature of levelling out.*

Translation study has always experienced a kind of difficulty with defining its subject. The issue of methods is even more critical for this branch of linguistics. How can one estimate equivalence or adequacy of translation in an impartial and dependable way? Are there any laws of translation process which influence its result? Linguists and translatologists have been studying these problems for the last fifty years.

In the nineties the field saw the first attempts to combine translation studies and corpus linguistics with its strict methods of analysing large text collections. Quantitative method allow to describe translation result in a detailed and reliable way and to come to some conclusions about the process itself. Review of research within this topic is presented, for example, in [Olohan, 2004].

It should not come as a surprise, that this inevitably lead translatologists to a disputable question: are there any objective statistical dependencies between source text and target text? This question is closely related to Baker's hypothesis about objective differences between any translated and non-translated texts [Baker, 1993]. If we solve this issue, we would be able to compare translations in an objective way; moreover, it would give additional clues to understand how human brain functions when we translate. Can we say that some statistical invariants of the text remain after translation or it is a kind of random choice? This is the topic of the present article.

Parallel corpus for preliminary research of this problem is: Kurt Vonnegut's novels «Cat's Cradle» (1963) and «Slaughterhouse-Five Or The Children's Crusade» (1969) and their Russian translations by Rita Rait-Kovaleva «Колыбель для кошки» and «Бойня номер пять или Крестовый поход детей» respectively. We took the texts from Moshkov e-library (http://www.lib.ru) and from the site «Vonnegut Books in English» (http://bg-studio.newmail.ru). The corpus was analysed with AntConc software (http://www.antlab.sci.waseda.ac.jp/software.html). Symbol case was not taken into account and we did no lemmatization.

In the source text of «Slaughter-House Five» there are 6247 word types and 50848 word tokens. Thus, the ratio is 0.12. Types used only once (*hapax legomena*) constitute 50% of word types.

In the source text of «Cat's Cradle» there are 6372 word types and 54353 word tokens. Thus, the ratio is also 0.12. 52% of word types are *hapax legomena*.

If we look at the frequency dictionary of both novels, we will see how the frequency is decreasing when the rank of a word type increases. This is the consequence of Zipf Law, which states that frequency of word types in any natural text is inversely proportional to their rank. It means that frequent word types are few and rare word types are many [Zipf, 1932].

The novels were split into equal fragments in order to find out whether type/token ratio coincides in different pieces of the texts. «Slaughter-House Five» was split into 10 fragments by chapters, and «Cat's Cradle» - into 11 fragments 300 lines each. Tables 1 and 2 show type/token ratios (TTR) in all the fragments:

*Table 1: Lexical statistical features of source text of «Slaughter-House Five»*

| Fragment | Number of word types | Number of word tokens | Types to tokens ratio (TTR[*]) |
|---|---|---|---|

* Types to tokens ratio

| | | | |
|---|---|---|---|
| 1 | 1411 | 5494 | 0,26 |
| 2 | 1805 | 7328 | 0,25 |
| 3 | 1376 | 4663 | 0,3 |
| 4 | 1099 | 3536 | 0,31 |
| 5 | 2389 | 11217 | 0,21 |
| 6 | 1207 | 4357 | 0,28 |
| 7 | 620 | 1676 | 0,37 |
| 8 | 1300 | 4686 | 0,28 |
| 9 | 1651 | 6529 | 0,25 |
| 10 | 523 | 1240 | 0,42 |
| The whole novel | 6247 | 50848 | 0,12 |

*Table 2: Lexical statistical features of source text of «Cat's Cradle»*

| Fragment | Number of word types | Number of word tokens | Types to tokens ratio (TTR) |
|---|---|---|---|
| 1 | 1479 | 6939 | 0,21 |
| 2 | 1236 | 4908 | 0,25 |
| 3 | 1514 | 5892 | 0,26 |
| 4 | 1373 | 5026 | 0,27 |
| 5 | 1483 | 5149 | 0,29 |
| 6 | 1167 | 4657 | 0,25 |
| 7 | 1326 | 4877 | 0,27 |
| 8 | 1081 | 3946 | 0,27 |
| 9 | 1352 | 5121 | 0,26 |
| 10 | 1615 | 5953 | 0,27 |
| 11 | 685 | 1885 | 0,36 |
| The whole novel | 6372 | 54353 | 0,12 |

It can be seen that while the length of fragments increases the number of word types does not increase in the same way - it happens much slower. This is additionally proved by general features of the novels: TTR for the whole novel in both cases is much lower than in any fragment. It should be noted that total TTR is equal for both novels. We can suppose that it is invariant for English fiction.

The texts possess not only straight laws like Zipf's one, but also some other, dynamic laws. Particularly, distribution of word frequencies depends on the text length and on the position that "we are now at", on how far we "have moved" from the beginning of the text to its end. This is called Heaps' law; according to it, the number of word types grows slower than the number of word tokens. Dictionary size is a sub-linear function of text size.

Speed of TTR decreasing (i.e., decreasing of probability that a new word type will be used) when a text develops is specific for particular languages, genres and/or authors [Bernhardsson et al, 2009]. Thus, how fast the angle of this ratio chart changes reflects peculiarities of the text. In the table 3 we show the development of TTR change in «Slaughter-House Five» if we consecutively measure it while the text grows, and figure 1 shows this development in a graphic way.

*Table 3: Development of TTR change in the source text of «Slaughter-House Five»*

| Fragments | Word types | Word tokens | TTR |
|---|---|---|---|
| 1 | 1411 | 5494 | 0,26 |
| 1+2 | 2640 | 12925 | 0,2 |
| 1+2+3 | 3300 | 17588 | 0,19 |
| 1+2+3+4 | 3742 | 21125 | 0,18 |
| 1+2+3+4+5 | 4849 | 32341 | 0,15 |
| 1+2+3+4+5+6 | 5187 | 36697 | 0,14 |
| 1+2+3+4+5+6+7 | 5320 | 38373 | 0,14 |
| 1+2+3+4+5+6+7+8 | 5700 | 43059 | 0,13 |
| 1+2+3+4+5+6+7+8+9 | 6156 | 49589 | 0,12 |
| The whole novel | 6247 | 50848 | 0,12 |

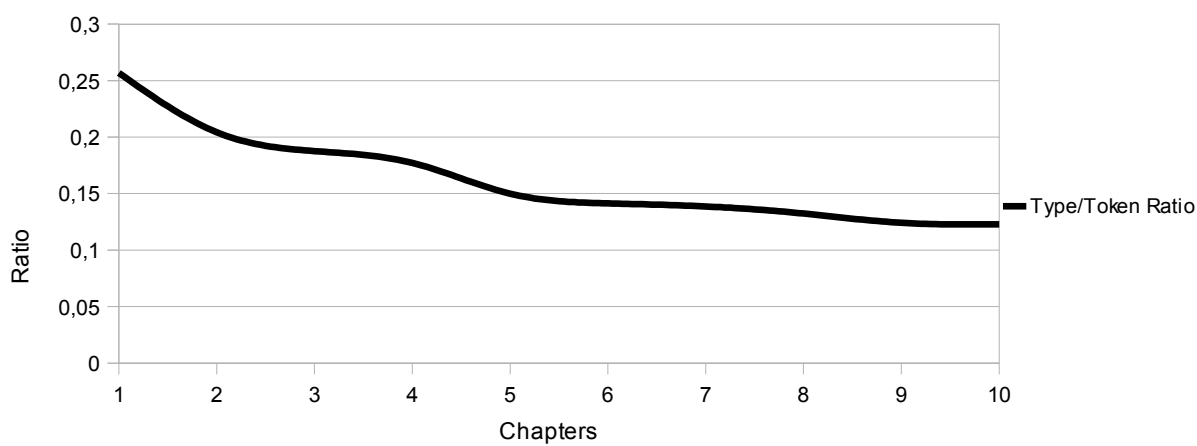

*Figure 1: Development of TTR in the source text of «Slaughter-House Five»*

Thus, we see smooth decrease of word types share from 25% of all the tokens in the first chapter down to 12% in the end of the novel. The notion of constant decrease of probability of using a new word is additionally supported by the

development of TTR for «Cat's Cradle» (figure 2):

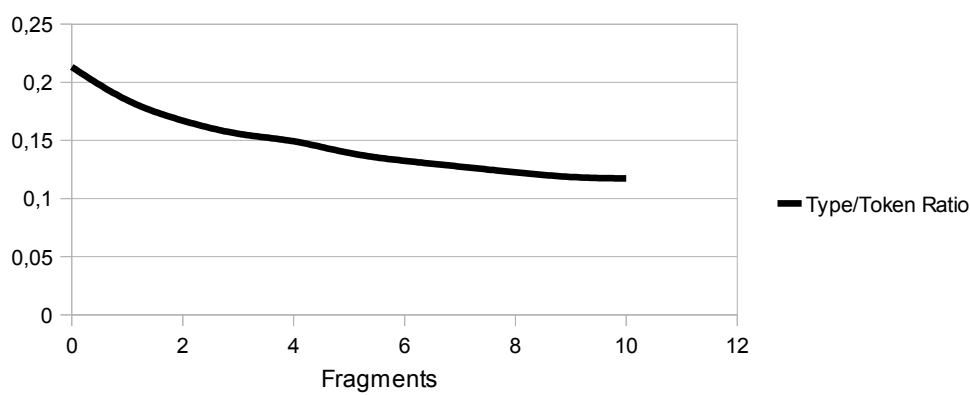

*Figure 2: Development of TTR in the source text of «Cat's Cradle»*

Let's hypothesize that the function, that links growth of word types number and word tokens number, is invariant for translation. We would be able to check this assumption if we calculate relevant parameters for Russian translations of «Slaughterhouse-Five» and «Cat's Cradle».

In Rait-Kovaleva's translation of "Slaughterhouse-Five" there are 12040 word types and 41596 word tokens. Thus, their ratio is 0.29. The number of *hapax legomena* is 8188, thus 68% among all word types.

In the translation of "Cat's Cradle" by the same translator there are 12070 word types and 44946 word tokens. Their ratio is 0.27. The number of *hapax legomena* is 8153, thus also 68% among all word types.

It should be noted that Russian texts also obey to Zipf Law, and beyond this, the target texts are very similar in lexical and statistical parameters.

It is rather interesting that Russian translations are almost 20% shorter than the source texts (it means they contain 20% less word tokens). It seems that such a ratio is characteristic for all English-Russian translations (cf. [Mikhailov, 2003]). This contradicts Nida&Taber hypothesis that good translated text is always longer than its source. At the same time, the translations under our scrutiny are definitely "good". It is proved by their large-scale popularity and Vonnegut's high appraisal of Rait-Kovaleva's professional skills (cf. his article *Invite Rita Rait to America!*). The issue of translation length is extensively covered in the above mentioned article by Mikhailov. Meanwhile, we get back to lexical and statistical features of the texts.

The fact that translated texts have two times more word types is caused by synthetic character of Russian language - the majority of words possess several morphological variants. At the same time corpus analysis tool considers "Джон" and "Джона" as two different word types (in fact one of them is just another in the genitive case). This influenced types/tokens ration which equals to 0.29 in the "Slaughterhouse..." translation and 0.27 in the "Cradle..." translation, as compared to 0.12 in both sources.

Initially, it seems that lexical statistical features of translated texts significantly differ from those of source texts: word types number is greater, word tokens number is vice versa lesser, their ratio is different, etc. Let's prove these differences with statistical methods.

We compiled table 4 for types/tokens ratios in each chapter of "Slaughterhouse..." translation, similar to the one we compiled for the source text.

*Table 4: Lexical statistical features of fragments of "Slaughterhouse..." translated text*

| Fragments | Word types | Word tokens | TTR |
|---|---|---|---|
| 1 | 2053 | 4513 | 0,45 |
| 2 | 2645 | 6002 | 0,44 |
| 3 | 1804 | 3673 | 0,49 |
| 4 | 1501 | 2908 | 0,52 |
| 5 | 3806 | 9315 | 0,41 |
| 6 | 1698 | 3508 | 0,48 |
| 7 | 730 | 1290 | 0,57 |
| 8 | 1851 | 3919 | 0,47 |
| 9 | 2405 | 5364 | 0,45 |
| 10 | 617 | 1004 | 0,61 |
| The whole novel | 12040 | 41596 | 0,29 |

It is obvious, that though absolute types/tokens ratios are greater than in the source texts, the relative values of TTR in separate fragments remained intact. The same is true for TTR development in the source and translation of "Cat's Cradle".

Let's check Heaps law behaviour in the translations. Table 5 presents development of TTR in translation of "Cat's Cradle" (translation of "Slaughterhouse..." shows similar features).

*Table 5: Development of TTR change in the translated text of «Cat's Cradle»*

| Fragments | Word types | Word tokens | TTR |
|---|---|---|---|
| 1 | 2511 | 6277 | 0,4 |
| 1+2 | 3845 | 10763 | 0,36 |
| 1+2+3 | 5342 | 15958 | 0,33 |
| 1+2+3+4 | 6608 | 20392 | 0,32 |
| 1+2+3+4+5 | 7647 | 24348 | 0,31 |
| 1+2+3+4+5+6 | 8604 | 28463 | 0,3 |
| 1+2+3+4+5+6+7 | 9309 | 31948 | 0,29 |
| 1+2+3+4+5+6+7+8 | 9871 | 34728 | 0,28 |
| 1+2+3+4+5+6+7+8+9 | 10852 | 39484 | 0,27 |
| 1+2+3+4+5+6+7+8+9+10 | 11919 | 44118 | 0,27 |
| The whole novel | 12070 | 44946 | 0,27 |

Word types number continues to grow slower than the number of word tokens. Are there any objective features of this "sluggishness", which survived translation?

Figure 3 presents (within one chart) diagrams of TTR development in the source and translated texts of «Slaughter-House Five» novel.

As we can see, translation TTR graph (black line) is next to copying source TTR graph (grey line), being only 0.2 points above. In fact, correlation analysis gives 99% probability of correlation between these number rows. The same is true for "Cat's Cradle".

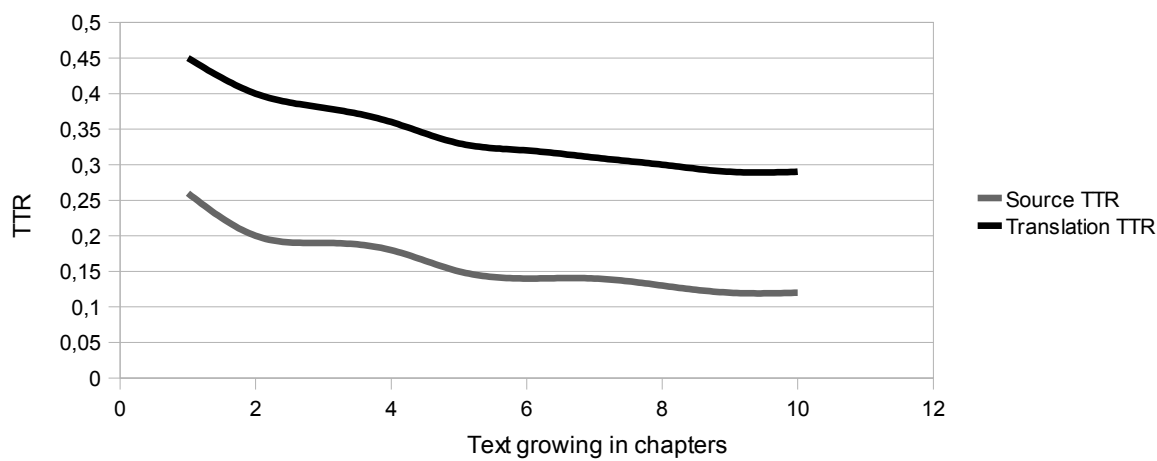

*Figure 3: Types/tokens ratio for source and translation of «Slaughter-House Five»*

The above numbers urged us to suppose a linear dependency between source TTR and translation TTR. We could even hypothesize that this correlation is a kind of dynamic invariance for given text and translator. However, if we compare source text of one novel and translated text of another, we find the same 99% correlation. It means we do not have sufficient grounds to state that there are some special features in a given pair "source-translation". We can only conclude that there is a certain correlation between TTR development in English fiction texts and their Russian translations, but it is not specific for separate works.

It is interesting to analyse in detail the changes in TTR dynamics, which were introduced in the course of translation. Also, it is possible to find any differences in this dynamics between source texts of both novels. Figure 4 presents an example, which proves that TTR diagrams in the source and translated texts to some extent differ. What exactly has changed in the statistical features of the texts?

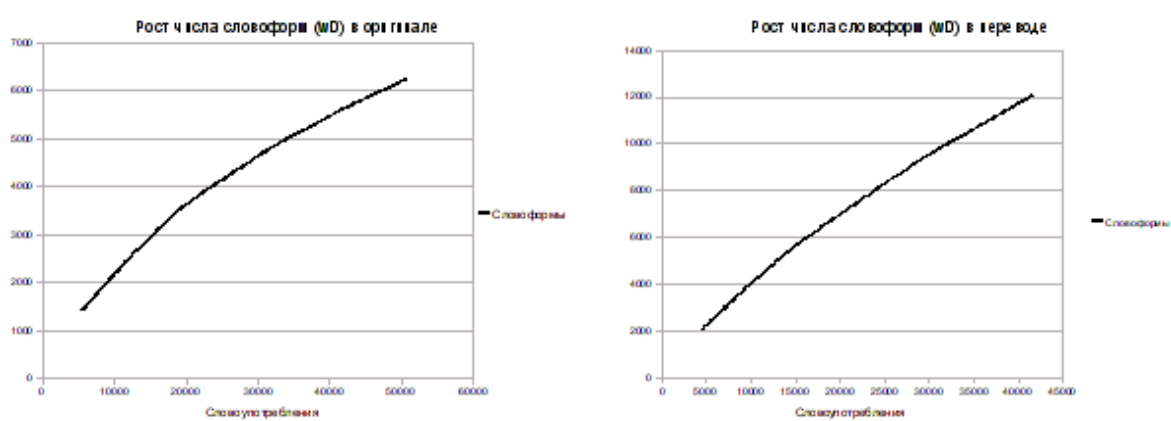

*Figure 4: Word types growing in the source and translated texts of Slaughter-House Five»*

Let's put word types number as *wD* (Distinct words), and word tokens number as *wγ*. In general, word types number is a function of word tokens number. According to Heaps Law, probability that a given word will occur in a text a given number of times changes with the development of the text. It means that the graph of *wD* is becoming more and more sloping, while *wγ* grows. This curve change is described by the following expression: *Pwγ(1)*. Thus, the probability that the next word will be unique (used only once) depends on *wγ*. *wD* dependency on *wγ* is a regular power equation $wD = a*(w\gamma)^b$, where *a* and *b* are coefficients specific for given language, text or author; *b* is always lesser than one [Serrano et al, 2009].

As TTR is defined by a power function, we are able to draw curves of power regression and calculate the values of this function coefficients in source and translation (that is, to obtain from experimental data the function which probably produced them). Figure 5 presents TTR diagrams in the source and translated text of "Cat's Cradle" and the corresponding curves of power regression (trends) and functions:

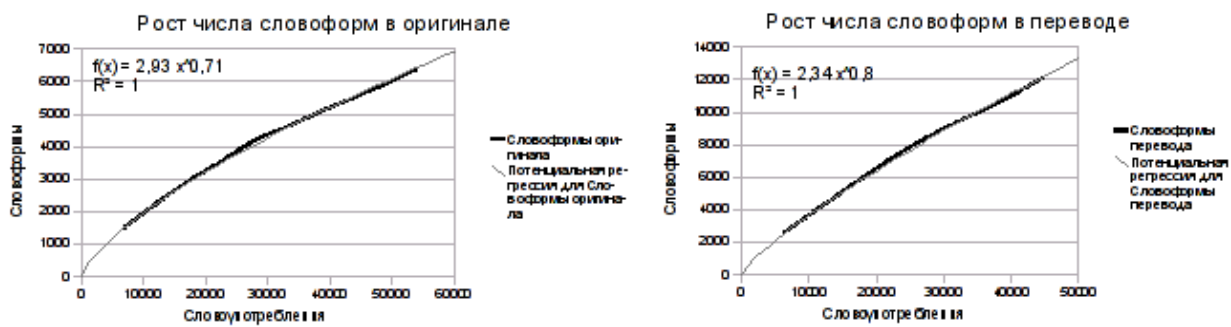

*Figure 5: Grow of word types number in the source and translated text of "Cat's Cradle"*

We can see that power regression gave us $wD = 2{,}93*(w\gamma)^{0,71}$ for the source text and $wD = 2{,}34*(w\gamma)^{0,8}$ for the translated text of «Cat's Cradle». For «Slaughter-House Five» the functions are $wD = 4{,}98*(w\gamma)^{0,66}$ and $wD = 2{,}72*(w\gamma)^{0,79}$ correspondingly.

It did not come as a surprise, that in general the function that binds together types number and tokens number really remains intact in the course of translation. It closely follows Heaps Law. But the coefficients of the equation change.

The changes touched both the exponent *b* (in both novels after the translation it increased) and the factor *a* (in both novels it decreased). Perhaps, the exponent has changed because of synthetic character of Russian language, where the probability of an author using a new word type is higher than in analytical English language. That's why the exponent increases and that means increasing the number of word types related to the number of word tokens. Thus, most likely the change of exponent is typical for "English-Russian" language pair.

It is still not clear why *a* factor has decreased. It contradicts the above mentioned tendency of Russian TTR to be higher than English one. Perhaps this is a random fluctuation, but it is reproduced in two translations. We can suppose that there exists some influence of so called "translator voice" (her individual features), but further research is needed to check this hypothesis, employing other translations by R. Rait-Kovaleva.

Considering translation universalia, another feature of the texts under our scrutiny should be noted. Coefficients of Heaps Law in translations of both novels are much closer to each other than in the source texts. E.g., source exponents are 0.66 and 0.71, while translations exponents are 0.79 and 0.80 correspondingly. This tendency is even more clear for *a* factor. In the source texts its values are 4.98 and 2.93, while in translations they are 2.72 and 2.34 correspondingly. It means that the original Vonnegut's novels are more diverse in the aspect of TTR, than Rait-Kovaleva's translations, where values are less scattered.

This indirectly proves the existence of one of translation universalia, formulated by M. Baker, namely "levelling out". Baker herself defines levelling out as "the tendency of translated text to gravitate around the centre of any continuum" [Baker, 1996]. This translation feature is considered to be the most difficult to verify and there is still few empiric grounds to prove or disprove its existence (it is even more true for non-European languages). That's why less variance of Vonnegut's translations in comparison with source texts (if only for two novels) can be an important step towards research in translation universalia.

Thus, there are some objective statistical differences between source and translated texts. It is possible to study them and to reveal linguistic laws responsible for their appearance in the course of translation. In particular, Russian translations of English fiction tend to level out (as in Baker), which is expressed in the dynamics of types to tokens ratio. On the other hand, general equation which binds these values (Heaps Law) remains intact after translation.

Revealing such dependencies allows for statistical comparing of different translations (or translators) and opens new directions for further research in translation as a type of human activity.